\documentclass[conference]{IEEEtran}
\IEEEoverridecommandlockouts
\usepackage{cite}
\usepackage{amsmath,amssymb,amsfonts}
\usepackage{graphicx}
\usepackage{textcomp}
\usepackage{xcolor}
\usepackage{algorithm}
\usepackage{algpseudocode}
\usepackage{booktabs}
\usepackage{mathabx}

\def\BibTeX{{\rm B\kern-.05em{\sc i\kern-.025em b}\kern-.08em
    T\kern-.1667em\lower.7ex\hbox{E}\kern-.125emX}}

\begin{document}

\title{Discriminative Deep Feature Visualization for Explainable Face Recognition
\thanks{$^\ast$Equal contribution}
\thanks{Support from XAIface CHIST-ERA-19-XAI-011 and the Swiss National Science Foundation (SNSF) 20CH21\_195532 is acknowledged.}
}
\author{\IEEEauthorblockN{Zewei Xu$^\ast$, Yuhang Lu$^\ast$, and Touradj Ebrahimi} 
\IEEEauthorblockA{\textit{Multimedia Signal Processing Group (MMSPG)} \\
\textit{\'Ecole Polytechnique F\'ed\'erale de Lausanne (EPFL)}\\
CH-1015 Lausanne, Switzerland \\
Email: firstname.lastname@epfl.ch}}

\maketitle

\begin{abstract}

Despite the huge success of deep convolutional neural networks in face recognition (FR) tasks, current methods lack explainability for their predictions because of their ``black-box'' nature. In recent years, studies have been carried out to give an interpretation of the decision of a deep FR system. However, the affinity between the input facial image and the extracted deep features has not been explored. 
This paper contributes to the problem of explainable face recognition by first conceiving a face reconstruction-based explanation module, which reveals the correspondence between the deep feature and the facial regions. To further interpret the decision of an FR model, a novel visual saliency explanation algorithm has been proposed. It provides insightful explanation by producing visual saliency maps that represent similar and dissimilar regions between input faces. A detailed analysis has been presented for the generated visual explanation to show the effectiveness of the proposed method. 

\end{abstract}

\begin{IEEEkeywords}
face recognition, explainable AI, deep learning
\end{IEEEkeywords}

\section{Introduction}

With the rise of deep learning, remarkable progress has been made in various computer vision applications, such as image classification and face recognition. In the past years, deep convolutional neural networks (DCNNs) have played a fundamental role in these tasks and significantly boosted their performance, even achieving higher accuracy than human observers \cite{triplet_ft}.
Despite the huge success, the decision made by these neural networks tends to be challenging to understand and interpret due to their ``black-box'' nature \cite{xbio,xAI_bg}.
Currently, deep learning-based technologies have been employed in multiple safety and security crucial domains. For example, deep face recognition techniques are widely used in access control systems, where a false positive prediction made by the ``black-box'' model can lead to severe consequences. 

Therefore, the general necessity of transparency and interpretability of deep learning techniques has inspired fast progress in the field of explainable artificial intelligence (XAI). Specifically, the explanation method should interpret the prediction of DCNNs in a precise and reliable manner in order to better comprehend the reasoning and potential vulnerability behind the decision system \cite{4principles,xexplanations}. 
In particular, this paper focuses on the problem of explainable face recognition (XFR). XFR generally studies the decision-making process of a deep face recognition system, i.e. how the model matches a given facial image over another \cite{xbio,XFR}. In this context, the explanation is usually presented in the form of attention or saliency maps for visualizing the discriminative areas on the faces.

In recent years, several approaches have been introduced to increase the explainability of face recognition techniques, allowing for further improvements in the reliability and even performance of the system. 
For example, in \cite{minplus}, perturbations were applied to input images and their impact on the output was analyzed. In \cite{hiding}, the contrastive excitation backpropagation method was used to identify the relevant regions of the images. In \cite{xcos}, a learnable module was plugged into existing face recognition models to enhance the overall explainability.
 
However, current popular XFR methods have shown some common limitations. First, the connection between the deep face representation and the discriminative facial regions has not been sufficiently explored in previous studies. In general, the decision-making process of a face recognition model fully relies on the similarity comparison between two or more deep feature vectors extracted by the FR model. 
In this context, the distances between specific dimensions of two features are expected to accurately reflect the discriminative regions on the input face images. 
This paper proposes a novel explanation approach based on face reconstruction, which reproduces a face image from its deep representation. Thereafter, any modification on the deep feature will reflect on the reconstructed image through forward propagation. 
Secondly, most of the current explanation methods focus on generating a saliency map that describes the similar regions between two given face images. Nevertheless, the dissimilarities between the inputs, particularly the non-matching face pairs, can also dominate the decision-making process. In this work, the proposed method is capable of providing richer explanation by producing saliency maps that respectively represent similar and dissimilar regions. 
In summary, the main contributions of the paper are as follows.
\begin{itemize}
    \item A new explainable face recognition framework is proposed, which explores the connection between facial images and deep features via a face reconstruction module. 
    \item A saliency map generation algorithm has been conceived, which provides explanation by highlighting the similarity and dissimilarity regions between two face images. 
    \item Extensive experimental results have shown that the proposed method can significantly enhance the model's explainability without affecting recognition performance.
\end{itemize}

\section{Related Works}

\subsection{Explainable Artificial Intelligence}
Explainable artificial intelligence refers to the problem of comprehending the predictions of a generic machine-learning model. Over the past years, the increasing demand for XAI technologies has promoted the development of numerous explanation methods that are based on different mechanisms. 
  
One major category of methods is based on backpropagation, which makes use of gradient information to directly identify the relevant pixels on the input image. 
Gradient \cite{gradient} evaluated the importance of the pixels by approximating the model with the gradient of the output with respect to the input. SmoothGrad \cite{smoothgrad} added noise to the input and averaged the gradients corresponding to multiple noisy images to obtain a de-noised saliency map. The Integrated Gradients approach \cite{integratedgrad} defined a baseline image and a straight-line path between the baseline and the input, and then performed integration of gradient over the path to estimate the relevance of each pixel.

Certain methods \cite{cam, gradcam, gradcam++, ecb, maxpool} additionally require accessing the intrinsic architecture or gradient information to compute the discriminative regions of the image. For example, CAM \cite{cam} modified the last layer of the deep model to perform an explanation, while its successors GradCAM \cite{gradcam} and GradCAM++ \cite{gradcam++} leveraged the internal gradients to retrieve the areas of interest from the deep features.

Another type of method works independently from the internal status of the deep model, which often refers to ``black-box'' explanation. LIME \cite{lime} analyzed the relation between the input image and the prediction in a perturbation-based manner. RISE \cite{rise} and its variant D-RISE \cite{drise} applied random masks to the input images and estimated the model's behavior by analyzing the impact of the perturbations on the posterior probability at the output.
On the other hand, approaches such as GAIN \cite{gain} integrated learnable modules into the model training process, providing attention maps as an explanation.

\subsection{Explainable Face Recognition}
While most of the classic explanation approaches were originally designed for image classification or detection tasks, various studies have been carried out recently to explain deep learning-based face recognition models. 

Most explanation methods in face recognition are represented through visualization of saliency maps. Earlier work \cite{hiding} first adapted several explanation methods from classification tasks, namely Grad-CAM \cite{gradcam}, Guided Grad-CAM \cite{guidedgradcam}, Gradient \cite{gradient}, etc., to face recognition and further benchmarked the explainability of the produced saliency maps. Similarly, \cite{XFR} proposed a Subtree Excitation Backprop (EBP) to generate an explainable saliency map, and meanwhile introduced an evaluation approach based on triplets along with an inpainting game protocol. 
More recently, Mery \cite{minplus,minplusold} has proposed several perturbation-based approaches to explain face verification algorithms without accessing the model. The produced saliency maps represent the most similar regions between the input pair during the recognition process. xFace \cite{xFace} achieved better performance by applying systematic occlusions to inputs and measuring the feature distance deviations.

Certain methods aim to increase explainability by introducing additional modules in the face recognition system. xCos \cite{xcos} designed a learnable module that was incorporated into a face recognition model to compute local similarities and produce spatial explanation. \cite{fad} proposed two loss functions and learned an interpretable face representation where each dimension of it depicts part of the face structure. 

Although many explainable face recognition methods manage to achieve compelling visual results, they \cite{lime,minplus,minplusold,lu2023towards, xFace} intend to explain the FR system in a ``black-box'' manner. There is a lack of discussion on the connection between the deep feature representation and the final decision of the FR model. Among prior art, \cite{maxpool, xFace} are closest to our proposed method.
The former added a MaxPooling layer to the end of the entire CNN architecture to decompose the final probability score. The latter explained a deep FR model by analyzing the similar and dissimilar regions between two faces. 
In this work, our method provides a more precise explanation by directly mapping the deep features to specific facial regions via a face reconstruction module.

\begin{figure*}[t]
\centerline{\includegraphics[width=0.92\linewidth]{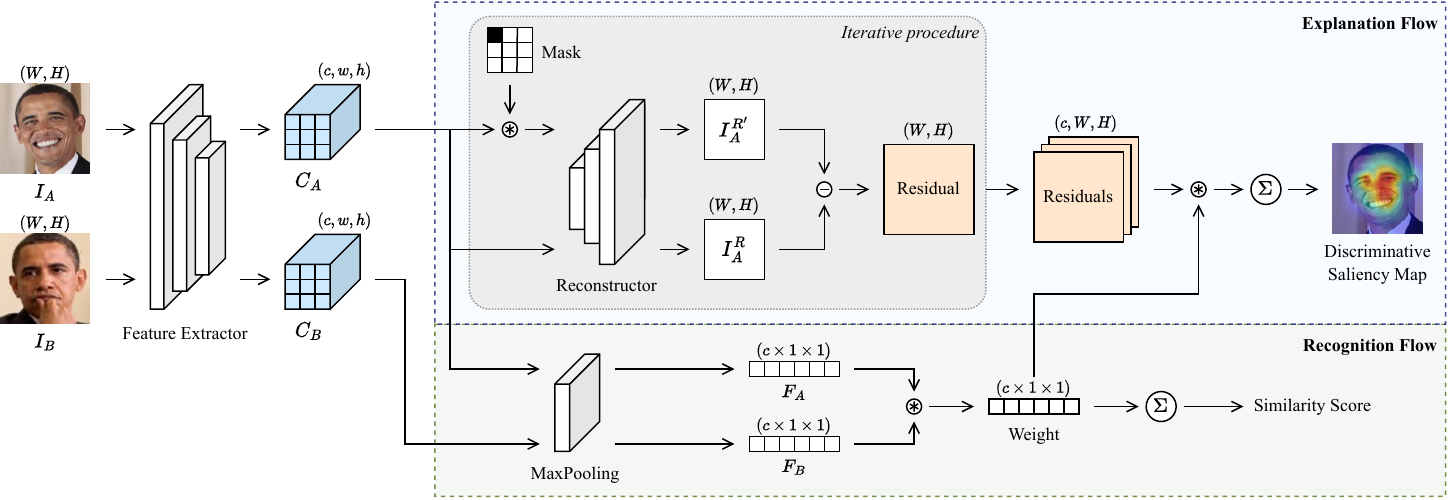}}
\caption{Two-stream workflow of the proposed explainable face recognition framework. The recognition flow performs normal face recognition tasks, while the explanation flow provides interpretations for the model's decision.}
\label{fig}
\label{fig:general_flow}
\end{figure*}

\section{Proposed Method}

Explanation methods for general deep learning-based computer vision systems are expected to provide a precise visual interpretation of the deep models' decisions. 
However, face recognition is different from other tasks due to an essential difference in the decision-making process, which often involves a comparison between two or more facial images. 
This paper defines the solution to explainable face recognition by answering the following questions: \textit{Which regions of two given faces are discriminative for the deep face recognition model to make predictions, and more specifically, which regions are the most similar or dissimilar to the deep model? }

\subsection{Feature Guided Explanation based on Face Reconstruction}

Previous research has explored the XFR problem from different perspectives but suffers from  drawbacks mainly in terms of stable outcomes. For example, gradient-based methods \cite{gradient,smoothgrad,integratedgrad} can be easily affected during backpropagation. ``Black-box'' explanation approaches based on perturbations \cite{rise,drise} would introduce inevitable randomness to the explanation map due to lack of a rigorous strategy to inject perturbation. 

One fact has been neglected that deep face recognition systems make decisions by directly comparing the distance between deep representations, where the most discriminative feature channels dominate the final prediction. Thus, this work explores the direct affinity between the specific discriminative feature channels and the input facial image in a forward propagation manner by leveraging a face reconstruction module. 
To the best of our knowledge, this is the first research that generates saliency map explanation by propagating the critical deep features to a reconstructed face image. 

{
\centering
\begin{minipage}{\linewidth}
\begin{algorithm}[H]
\caption{}
\begin{algorithmic}
\State \textbf{Input:}
\State Face images: $I_A$, $I_B$
\State Number of feature channels: $c$
\State Threshold: $T$
\State Feature extractor: \texttt{f()}
\State Face reconstructor: \texttt{Recon()}
\vspace{0.2em}
\hrule
\vspace{0.2em}
    \State $C_A, F_A \gets \texttt{f}(I_A)$
    \State $C_B, F_B \gets \texttt{f}(I_B)$  
    \State $F_A, F_B \gets \texttt{normalization}(F_A, F_B)$
    \State $weights \gets F_A \oasterisk F_B - T$
    \State $I_A^R \gets \texttt{Recon}(C_A)$
    \State $C_A^\prime \gets C_A$
    \For{$i=1:c$}
        \State $max(C^\prime_A[i]) \gets 0$
        \State $I_A^R, I_A^{R\prime} \gets \texttt{Recon}(C_A), \texttt{Recon}(C^\prime_A)$
	  \State $Residual[i] \gets abs(I_A^R - I_A^{R\prime})$
        \State $Residual[i] \gets \texttt{normalization}(Residual[i])$
    \EndFor
    \State $H \gets \sum Residual \oasterisk weights$
    \State $S \gets abs(H)$  \algorithmiccomment{discriminative map}
    \State $S^+ \gets ReLU(H)$ \algorithmiccomment{similarity map}
    \State $S^- \gets ReLU(-H)$ \algorithmiccomment{dissimilarity map}
\hrule
\\
\vspace{0.2em}
\Return $S, S^+, S^-$
\end{algorithmic}
\label{alg:1}
\end{algorithm}
\end{minipage}
\par
}
\vspace{0.5em}

The overall explainable face recognition framework is illustrated in Fig. \ref{fig:general_flow}, which comprises two separate streams, i.e. the recognition flow and the explanation flow. The former performs standard face recognition tasks. In the explanation workflow, a face reconstruction network is introduced with the goal of reproducing a facial image from its deep convolutional representations. 
The main idea is that any modification on the deep feature will reflect on the rebuilt images via the reconstruction network. 
Moreover, the difference between the unmodified rebuilt image and the feature-masked reconstructed image can indicate the specific face region that corresponds to the masked channel in the feature vector.

\subsection{Saliency Map Generation}

The face reconstruction-based explanation flow provides means of understanding the deep feature representation, whilst it cannot yet interpret the predictions of a face recognition system. 
This section proposes a novel explainable saliency map generation algorithm that both leverages the feature-guided face reconstruction workflow and takes into account the similarity calculation between two face pairs. 
Moreover, this work decomposes the produced discriminative saliency map into two visual representations, i.e. similarity and dissimilarity heatmaps, for a higher-level interpretation.  

The procedure of the saliency map generation method is described in both Fig. \ref{fig:general_flow} and Algorithm \ref{alg:1}. In general, the explanation process is designed in a channel-wise iterative manner. Given input image $I_A$, the convolutional feature map $C_A$ of size $(c, w, h)$ is first extracted and accessed. Then, the explanation module iterates along each channel of $C_A$ and modifies the maximum value of this channel by setting $max(C_{Ai})=0$. Afterward, the reconstruction module rebuilds face images $\{I_A^{R}, I_A^{R\prime}\}$ from both original and modified features, and the residual difference between them is calculated and normalized. After iteration, the stacked residual maps represent a channel-wise mapping between the convolutional feature map and the specific regions on the facial image.

At the same time, the standard face recognition flow calculates the cosine distance between two deep face representations. 
Equation \ref{eq:sim} shows the computation of the similarity score $s$, where the feature vectors $F_1, F_2$ are first normalized and then channel-wise multiplication is computed. 
The value of every single channel of the resulting vector represents a feature-level distance. Therefore, the entire vector will serve as a channel-wise importance weight for the aforementioned residual maps in order to combine the final saliency map.

\begin{equation} \label{eq:sim}
s(F_1, F_2) = \frac{F_1\cdot F_2}{||F_1|| \, ||F_2||} =\frac{\sum_{i=1}^c(f_{1i}*f_{2i})}{||F_1|| \, ||F_2||}.
\end{equation}

Moreover, the proposed method is capable of further decomposing the discriminative saliency map into similarity and dissimilarity maps through analysis of the weight vector. Intuitively, the large and positive values represent the similar feature channels between two face representations, while the small and negative ones represent dissimilar channels. As explained in Algorithm \ref{alg:1}, the weight vector is re-balanced by subtracting the decision threshold, and the desired similarity and dissimilarity maps are obtained by simply separating the discriminative saliency map. 

\subsection{Model Architecture and Loss Function}
As depicted in Fig. \ref{fig:general_flow}, the proposed explanation framework adopts a two-stream workflow that allows for providing recognition predictions and decision interpretation in a precise and efficient manner. This section describes in detail the network architecture and training process. 

In the recognition flow, a minor modification has been introduced to the conventional face recognition pipeline. Specifically, it removes the last fully connected layer in the ResNet backbone and replaces it with a global MaxPooling layer, which preserves the spatial information to establish a correspondence between the convolutional feature and the facial image. 
In the explanation flow, a face reconstruction network comprising a stack of four transposed convolutional layers is employed, which takes the convolutional features as input to rebuild a facial image. 

During the training phase, the recognition flow and the explanation flow, mainly referring to the reconstruction network, are jointly trained by optimizing both the ArcFace \cite{arcface} recognition loss $\mathcal{L}_{id}$ and an MSE loss $\mathcal{L}_{rec}$ between the original and reconstructed faces $\{I_A, I^R_A\}$. The overall loss function is defined as follows.
\begin{equation}
    \mathcal{L}=\mathcal{L}_{id}+\lambda \cdot \mathcal{L}_{rec} \, ,
\end{equation}
where $\lambda=1$ is selected for all the experiments in this paper. 

\section{Experiments}
\subsection{Implementation Details}

During the experiments, this work adopts the well-known ArcFace \cite{arcface} method as the face recognition pipeline, using ResNet-50 \cite{resnet50} as a feature extractor. 
The face recognition system and the reconstruction module are jointly trained on the MS1Mv2 dataset \cite{ms1m} for 25 epochs.
Specifically, the face recognition pipeline is trained with the SGD optimizer with an initial learning rate of 0.02, while the reconstruction network uses the Adam optimizer with an initial learning rate of $2\times 10^{-4}$.
During evaluation, the performance of the proposed method is tested on LFW \cite{lfw}, AgeDB-30 \cite{agedb}, CFP \cite{cfp}, IJB-B \cite{ijbb}, and IJB-C \cite{ijbc} datasets.

\subsection{Evaluation Methodology}


The evaluation for the proposed explainable face recognition framework includes three phases. 
First, the recognition performance is evaluated on five popular FR benchmarks and is compared with the original ArcFace pipeline.
Then, a visual demonstration of the generated explanation saliency maps is presented. In the third phase, a quantitative evaluation methodology called ``hiding game'' is employed for a fair comparison among current state-of-the-art saliency map-based explanation methods. This evaluation approach was first conceived in \cite{hiding}. They first sort the generated heatmap in ascending order and mask the least important pixels. The face recognition model is then tested with the obscured images. Ideally, the more accurate the saliency map, the higher the evaluation accuracy because only the most critical pixels are maintained. 
However, directly changing the pixel values will modify the original data distribution, which leads to less accurate results. 
Therefore, instead of simply masking the least important pixels, this paper proposes to blur the pixels with a Gaussian kernel, which is often considered an intuitive way of representing missing information. 


\subsection{Experimental Results}

\begin{table}[t]
\caption{Recognition performance of the explainable face recognition framework.}
\begin{center}
\resizebox{0.9\linewidth}{!}{
\begin{tabular}{l|c c c c c }
\hline
Model & LFW & AgeDB & CFP & IJB-B & IJB-C \\
\hline
ArcFace & 99.78 & 98.03 & 96.71 & \textbf{95.00} & \textbf{96.45} \\
\textbf{Ours} & \textbf{99.83} & \textbf{98.07} & \textbf{97.90} & 94.49 & 95.95 \\
\hline
\end{tabular}}
\label{tab:model_comparison}
\end{center}
\end{table}

\begin{figure*}[t]
\centerline{\includegraphics[width=0.95\linewidth]{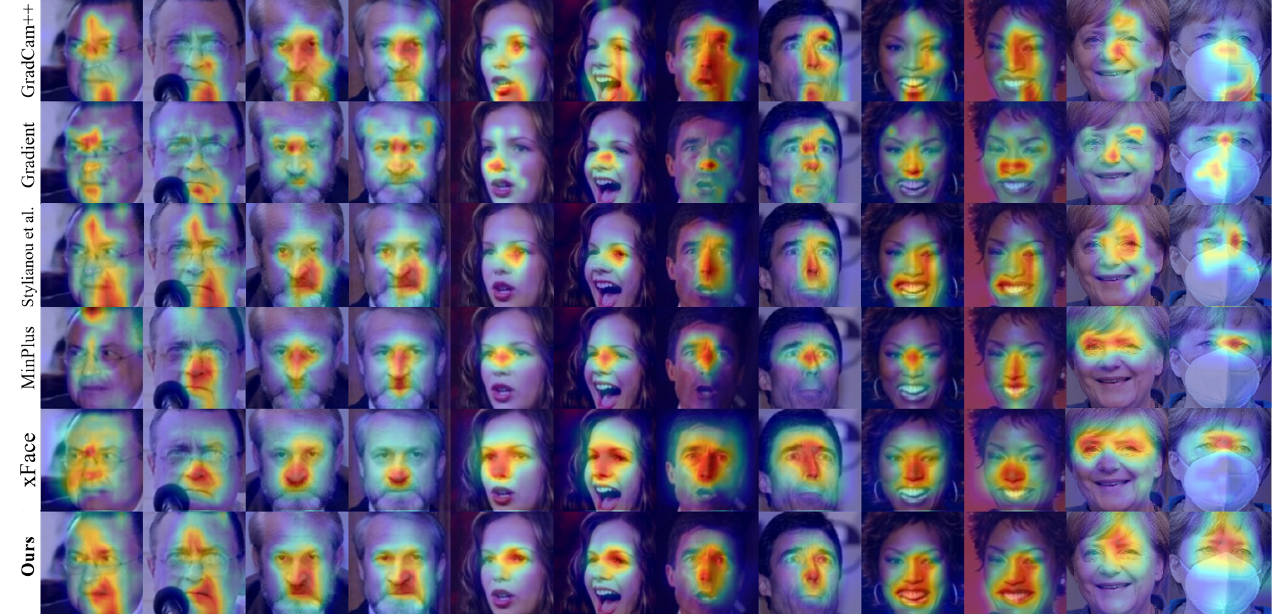}}
\caption{Visual comparison of saliency maps generated by six different explainability methods.}
\label{fig}
\label{fig:lfw_probe_mate}
\end{figure*}

\begin{figure}[t]
\centerline{\includegraphics[width=0.95\linewidth]{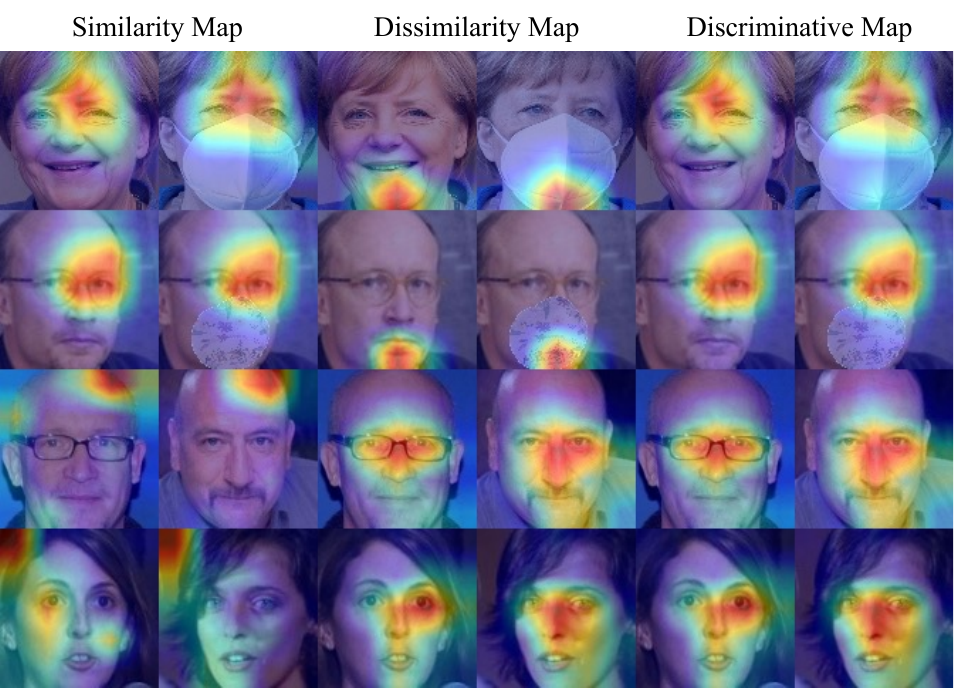}}
\caption{Saliency map explanations highlighting the facial regions that the deep model believes are similar, dissimilar, and discriminative.}
\label{fig}
\label{fig:local_data}
\end{figure}

\subsubsection{Recognition Performance}
The objective of this section is to show there is no performance deterioration after integrating the proposed reconstruction-based explanation module. The face verification task is performed on five different datasets. 
Table \ref{tab:model_comparison} shows that the explainable face recognition framework is capable of achieving similar accuracy to the original ArcFace pipeline on all the test datasets and even slightly outperforms it on LFW, AgeDB, and CFP, providing interpretation on the deep recognition system in the meantime.

\subsubsection{Visual Results}

In this section, the visual explanation for the deep face recognition system is presented through saliency maps. The experiments are conducted under the face verification scenario. 

Fig. \ref{fig:lfw_probe_mate} demonstrates a comparison between saliency maps produced by six explanation methods. The first three rows are from well-known XAI methods adapted to the XFR task, while the fourth and fifth rows are produced by state-of-the-art model-agnostic XFR methods. Every two columns represent a to-be-explained matching input pair and the discriminative regions in them highlighted by different methods. As a result, the proposed method is capable of producing meaningful explanation maps, which focus on critical facial characteristics, e.g. lips, nose, eyes, mouth, etc., and vary for different subjects. It is also notable that the proposed method spotlights discriminative regions in unusual cases, such as the opening mouth and teeth in the second last example and the non-masked regions in the last sample. In contrast, many other approaches cannot provide stable outcome in various test cases and fail in producing meaningful explanation in the last sample. For a more straightforward comparison, the quantitative evaluation is introduced in the next subsection.

The results in Fig. \ref{fig:local_data} offer a deeper analysis of how the proposed explanation method interprets the decision of a deep face recognition system by decomposing the saliency map into similarity and dissimilarity maps. 
The first two rows are examples of image pairs belonging to the same subjects but with mask occlusions. It is shown that the decomposed similarity map mainly highlights the facial regions while the dissimilarity map highlights the occluded parts. The discriminative saliency maps provide a final interpretation that the similarity map actually weighs more and the deep model manages to verify the occluded image based on the unmasked areas, e.g. eyes and forehead. 
Finally, the last two rows include two pairs of non-matching subjects. It is notable that the dissimilarity map concentrates on critical facial areas while the similarity map only focuses on the backgrounds. The discriminative maps show that the deep model relies more on the dissimilarity map to provide the final decision.

\begin{figure}[t]
\centerline{\includegraphics[width=0.9\linewidth]{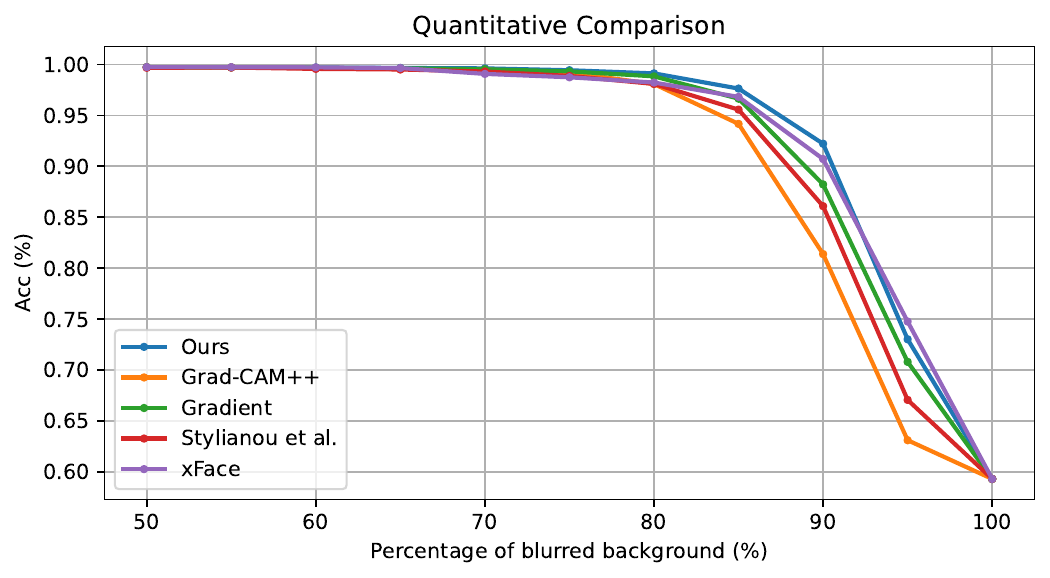}}
\caption{Quantitative performance evaluation on LFW dataset.}
\label{fig}
\label{fig:roc}
\end{figure}

\subsubsection{Quantitative Results}

This section gives quantitative evaluation results for a fair comparison among different explanation methods, namely GradCAM++ \cite{gradcam++}, Gradient \cite{gradient}, Stylianou et al. \cite{maxpool}, and xFace \cite{xFace}. The MinPlus \cite{minplus} method is not selected for this experiment because it cannot explain predictions on non-matching faces. 
Fig. \ref{fig:roc} depicts the results after applying the ``hiding game'' evaluation, which clearly shows that the saliency maps produced by the proposed method remain the most accurate when different percentages of the least important pixels are blurred out. 
Table~\ref{tab:hiding} reports the Area Under the Curve (AUC) metric for the Fig.~\ref{fig:roc} along with test results on AgeDB-30 \cite{agedb} and CFP-FP \cite{cfp} datasets under the same setting. It validates the conclusion that the proposed saliency maps are more accurate than the state-of-the-art in describing the most discriminative facial regions. 


\begin{table}[t]
\caption{Quantitative performance evaluation on LFW, AgeDB, and CFP datasets. AUC scores from the ``Hiding Game'' are reported.}
\begin{center}
\resizebox{0.43\textwidth}{!}{
\begin{tabular}{l|c c c}
\hline
Methods & LFW & AgeDB-30 & CFP-FP \\
\hline
GradCAM++ \cite{gradcam++} & 88.05 & 83.00 & 82.20 \\
Gradient \cite{gradient} & 89.68 & 85.13 & 84.24 \\
Stylianou et al. \cite{maxpool} & 88.90 & 83.73 & 81.99 \\
xFace \cite{xFace} & 90.14 & 85.82 & 82.96 \\ 
\textbf{Ours} & \textbf{90.39} & \textbf{86.23} & \textbf{85.26} \\
\hline
\end{tabular}}
\label{tab:hiding}
\end{center}
\end{table}

\section{Conclusion}
In this paper, a two-stream explainable face recognition framework is proposed. The connection between an input facial image and the corresponding extracted deep feature is analyzed in detail via a face reconstruction module. Furthermore, a saliency map generation algorithm has been conceived to provide visual interpretations for the prediction of the face recognition system. 
Both the visual and quantitative evaluation results have shown that the proposed method can significantly improve the interpretability of a face recognition system without affecting its recognition performance.

{
\bibliographystyle{IEEEtran}
\bibliography{IEEEabrv,mybib}

\begin{thebibliography}{10}
\providecommand{\url}[1]{#1}
\csname url@samestyle\endcsname
\providecommand{\newblock}{\relax}
\providecommand{\bibinfo}[2]{#2}
\providecommand{\BIBentrySTDinterwordspacing}{\spaceskip=0pt\relax}
\providecommand{\BIBentryALTinterwordstretchfactor}{4}
\providecommand{\BIBentryALTinterwordspacing}{\spaceskip=\fontdimen2\font plus
\BIBentryALTinterwordstretchfactor\fontdimen3\font minus
  \fontdimen4\font\relax}
\providecommand{\BIBforeignlanguage}[2]{{%
\expandafter\ifx\csname l@#1\endcsname\relax
\typeout{** WARNING: IEEEtran.bst: No hyphenation pattern has been}%
\typeout{** loaded for the language `#1'. Using the pattern for}%
\typeout{** the default language instead.}%
\else
\language=\csname l@#1\endcsname
\fi
#2}}
\providecommand{\BIBdecl}{\relax}
\BIBdecl

\bibitem{triplet_ft}
M.~Wang and W.~Deng, ``Deep face recognition: A survey,''
  \emph{Neurocomputing}, vol. 429, pp. 215--244, 2021.

\bibitem{xbio}
P.~C. Neto, T.~Gonçalves, J.~R. Pinto, W.~Silva, A.~F. Sequeira, A.~Ross, and
  J.~S. Cardoso, ``Explainable biometrics in the age of deep learning,''
  \emph{arXiv preprint arXiv:2208.09500}, 2022.

\bibitem{xAI_bg}
A.~Rai, ``Explainable ai: From black box to glass box,'' \emph{Journal of the
  Academy of Marketing Science}, vol.~48, pp. 137--141, 2020.

\bibitem{4principles}
P.~J. Phillips and M.~A. Przybocki, ``Four principles of explainable ai as
  applied to biometrics and facial forensic algorithms,'' \emph{arXiv preprint
  arXiv:2002.01014}, 2020.

\bibitem{xexplanations}
L.~H. Gilpin, D.~Bau, B.~Z. Yuan, A.~Bajwa, M.~A. Specter, and L.~Kagal,
  ``Explaining explanations: An approach to evaluating interpretability of
  machine learning,'' \emph{arXiv preprint arXiv:1806.00069}, p. 118, 2018.

\bibitem{XFR}
J.~R. Williford, B.~B. May, and J.~Byrne, ``Explainable face recognition,'' in
  \emph{Computer Vision--ECCV 2020: 16th European Conference, Glasgow, UK,
  August 23--28, 2020, Proceedings, Part XI}, 2020, pp. 248--263.

\bibitem{minplus}
D.~Mery, ``True black-box explanation in facial analysis,'' in
  \emph{Proceedings of the IEEE/CVF Conference on Computer Vision and Pattern
  Recognition}, 2022, pp. 1596--1605.

\bibitem{hiding}
G.~Castanon and J.~Byrne, ``Visualizing and quantifying discriminative features
  for face recognition,'' in \emph{2018 13th IEEE International Conference on
  Automatic Face \& Gesture Recognition (FG 2018)}.\hskip 1em plus 0.5em minus
  0.4em\relax IEEE, 2018, pp. 16--23.

\bibitem{xcos}
Y.~Lin, Z.~Y. Liu, Y.~Chen, Y.~Wang, H.~Lee, Y.~Chen, Y.~Chang, and W.~H. Hsu,
  ``xcos: An explainable cosine metric for face verification task,'' \emph{ACM
  Transactions on Multimedia Computing, Communications, and Applications
  (TOMM)}, vol.~17, no.~3s, pp. 1--16, 2021.

\bibitem{gradient}
K.~Simonyan, A.~Vedaldi, and A.~Zisserman, ``Deep inside convolutional
  networks: Visualising image classification models and saliency maps,''
  \emph{arXiv preprint arXiv:1312.6034}, 2013.

\bibitem{smoothgrad}
D.~Smilkov, N.~Thorat, B.~Kim, F.~B. Vi{\'{e}}gas, and M.~Wattenberg,
  ``Smoothgrad: removing noise by adding noise,'' \emph{arXiv preprint
  arXiv:1706.03825}, 2017.

\bibitem{integratedgrad}
M.~Sundararajan, A.~Taly, and Q.~Yan, ``Axiomatic attribution for deep
  networks,'' in \emph{International conference on machine learning}.\hskip 1em
  plus 0.5em minus 0.4em\relax PMLR, 2017, pp. 3319--3328.

\bibitem{cam}
B.~Zhou, A.~Khosla, {\`{A}}.~Lapedriza, A.~Oliva, and A.~Torralba, ``Learning
  deep features for discriminative localization,'' in \emph{Proceedings of the
  IEEE conference on computer vision and pattern recognition}, 2016, pp.
  2921--2929.

\bibitem{gradcam}
R.~R. Selvaraju, A.~Das, R.~Vedantam, M.~Cogswell, D.~Parikh, and D.~Batra,
  ``Grad-cam: Why did you say that?'' \emph{arXiv preprint arXiv:1611.07450},
  2016.

\bibitem{gradcam++}
A.~Chattopadhyay, A.~Sarkar, P.~Howlader, and V.~N. Balasubramanian,
  ``Grad-cam++: Generalized gradient-based visual explanations for deep
  convolutional networks,'' in \emph{2018 IEEE winter conference on
  applications of computer vision (WACV)}.\hskip 1em plus 0.5em minus
  0.4em\relax IEEE, 2018, pp. 839--847.

\bibitem{ecb}
J.~Zhang, Z.~Lin, J.~Brandt, X.~Shen, and S.~Sclaroff, ``Top-down neural
  attention by excitation backprop,'' \emph{International Journal of Computer
  Vision}, vol. 126, no.~10, pp. 1084--1102, 2018.

\bibitem{maxpool}
A.~Stylianou, R.~Souvenir, and R.~Pless, ``Visualizing deep similarity
  networks,'' in \emph{2019 IEEE winter conference on applications of computer
  vision (WACV)}.\hskip 1em plus 0.5em minus 0.4em\relax IEEE, 2019, pp.
  2029--2037.

\bibitem{lime}
M.~T. Ribeiro, S.~Singh, and C.~Guestrin, ``" why should i trust you?"
  explaining the predictions of any classifier,'' in \emph{Proceedings of the
  22nd ACM SIGKDD international conference on knowledge discovery and data
  mining}, 2016, pp. 1135--1144.

\bibitem{rise}
V.~Petsiuk, A.~Das, and K.~Saenko, ``Rise: Randomized input sampling for
  explanation of black-box models,'' \emph{arXiv preprint arXiv:1806.07421},
  2018.

\bibitem{drise}
V.~Petsiuk, R.~Jain, V.~Manjunatha, V.~I. Morariu, A.~Mehra, V.~Ordonez, and
  K.~Saenko, ``Black-box explanation of object detectors via saliency maps,''
  in \emph{Proceedings of the IEEE/CVF Conference on Computer Vision and
  Pattern Recognition}, 2021, pp. 11\,443--11\,452.

\bibitem{gain}
K.~Li, Z.~Wu, K.~Peng, J.~Ernst, and Y.~Fu, ``Tell me where to look: Guided
  attention inference network,'' in \emph{Proceedings of the IEEE conference on
  computer vision and pattern recognition}, 2018, pp. 9215--9223.

\bibitem{guidedgradcam}
R.~R. Selvaraju, M.~Cogswell, A.~Das, R.~Vedantam, D.~Parikh, and D.~Batra,
  ``Grad-cam: Visual explanations from deep networks via gradient-based
  localization,'' in \emph{Proceedings of the IEEE international conference on
  computer vision}, 2017, pp. 618--626.

\bibitem{minplusold}
D.~Mery and B.~Morris, ``On black-box explanation for face verification,'' in
  \emph{Proceedings of the IEEE/CVF Winter Conference on Applications of
  Computer Vision}, 2022, pp. 3418--3427.

\bibitem{xFace}
M.~Knoche, T.~Teepe, S.~H{\"o}rmann, and G.~Rigoll, ``Explainable
  model-agnostic similarity and confidence in face verification,'' in
  \emph{Proceedings of the IEEE/CVF Winter Conference on Applications of
  Computer Vision}, 2023, pp. 711--718.

\bibitem{fad}
B.~Yin, L.~Tran, H.~Li, X.~Shen, and X.~Liu, ``Towards interpretable face
  recognition,'' in \emph{Proceedings of the IEEE/CVF International Conference
  on Computer Vision}, 2019, pp. 9348--9357.

\bibitem{lu2023towards}
Y.~Lu, Z.~Xu, and T.~Ebrahimi, ``Towards visual saliency explanations of face
  verification,'' \emph{arXiv preprint arXiv:2305.08546}, 2023.

\bibitem{arcface}
J.~Deng, J.~Guo, N.~Xue, I.~Kotsia, and S.~Zafeiriou, ``Arcface: Additive
  angular margin loss for deep face recognition,'' in \emph{Proceedings of the
  IEEE/CVF conference on computer vision and pattern recognition}, 2019, pp.
  4690--4699.

\bibitem{resnet50}
K.~He, X.~Zhang, S.~Ren, and J.~Sun, ``Deep residual learning for image
  recognition,'' in \emph{Proceedings of the IEEE conference on computer vision
  and pattern recognition}, 2016, pp. 770--778.

\bibitem{ms1m}
Y.~Guo, L.~Zhang, Y.~Hu, X.~He, and J.~Gao, ``Ms-celeb-1m: A dataset and
  benchmark for large-scale face recognition,'' in \emph{Computer Vision--ECCV
  2016: 14th European Conference, Amsterdam, The Netherlands, October 11-14,
  2016, Proceedings, Part III 14}, 2016, pp. 87--102.

\bibitem{lfw}
G.~B. Huang, M.~Ramesh, T.~Berg, and E.~Learned-Miller, ``Labeled faces in the
  wild: A database forstudying face recognition in unconstrained
  environments,'' in \emph{Workshop on faces in'Real-Life'Images: detection,
  alignment, and recognition}, 2008.

\bibitem{agedb}
S.~Moschoglou, A.~Papaioannou, C.~Sagonas, J.~Deng, I.~Kotsia, and
  S.~Zafeiriou, ``Agedb: the first manually collected, in-the-wild age
  database,'' in \emph{proceedings of the IEEE conference on computer vision
  and pattern recognition workshops}, 2017, pp. 51--59.

\bibitem{cfp}
S.~Sengupta, J.~Chen, C.~Castillo, V.~M. Patel, R.~Chellappa, and D.~W. Jacobs,
  ``Frontal to profile face verification in the wild,'' in \emph{2016 IEEE
  winter conference on applications of computer vision (WACV)}.\hskip 1em plus
  0.5em minus 0.4em\relax IEEE, 2016, pp. 1--9.

\bibitem{ijbb}
C.~Whitelam, E.~Taborsky, A.~Blanton, B.~Maze, J.~Adams, T.~Miller, N.~Kalka,
  A.~K. Jain, J.~A. Duncan, K.~Allen, J.~Cheney, and P.~Grother, ``Iarpa janus
  benchmark-b face dataset,'' in \emph{proceedings of the IEEE conference on
  computer vision and pattern recognition workshops}, 2017, pp. 90--98.

\bibitem{ijbc}
B.~Maze, J.~Adams, J.~A. Duncan, N.~Kalka, T.~Miller, C.~Otto, A.~K. Jain,
  W.~T. Niggel, J.~Anderson, J.~Cheney, and P.~Grother, ``Iarpa janus
  benchmark-c: Face dataset and protocol,'' in \emph{2018 international
  conference on biometrics (ICB)}.\hskip 1em plus 0.5em minus 0.4em\relax IEEE,
  2018, pp. 158--165.

\end{thebibliography}
}

\end{document}